\documentclass[letterpaper]{article} 
\usepackage[preprint]{aaai2027}  
\usepackage[hyphens]{url}  
\usepackage{graphicx} 
\usepackage{natbib}  
\usepackage{caption} 
\usepackage{algorithm}
\usepackage{algorithmic}
\usepackage{booktabs}
\usepackage{rotating}
\usepackage{array}
\usepackage{tabularx}
\usepackage{makecell}
\usepackage{pifont} 
\usepackage{amssymb}
\usepackage{amsmath}
\usepackage{multirow}

\usepackage{newfloat}
\usepackage{listings}
\DeclareCaptionStyle{ruled}{labelfont=normalfont,labelsep=colon,strut=off} 
\floatstyle{ruled}
\newfloat{listing}{tb}{lst}{}
\floatname{listing}{Listing}
\title{PRIME: Predictive Routing of Interventions by Modeling Effects \\for Scientific Ideation}

\author{
    Fengxian Ji\textsuperscript{\rm 1,2,3}\equalcontrib,
    Xie Yu\textsuperscript{\rm 3}\equalcontrib,
    Zhexuan Cui\textsuperscript{\rm 3}\equalcontrib,
    Yuke Li\textsuperscript{\rm 3},
    Zirui Song\textsuperscript{\rm 2},
    Fan Zhang\textsuperscript{\rm 2},
    Qian Jiang\textsuperscript{\rm 3},\\
    Jingpu Yang\textsuperscript{\rm 3,4},
    Xiuying Chen\textsuperscript{\rm 2},
    Preslav Nakov\textsuperscript{\rm 2},
    Zhuohan Xie\textsuperscript{\rm 2}\corresponding
}
\affiliations{
    \textsuperscript{\rm 1}School of Artificial Intelligence, Wuhan University\quad
    \textsuperscript{\rm 2}MBZUAI\\
    \textsuperscript{\rm 3}Northeastern University\quad
    \textsuperscript{\rm 4}Zhongguancun Academy\\
    jingpuyang290@gmail.com,
    fengxian.ji@mbzuai.ac.ae, fan.zhang@mbzuai.ac.ae, zhuohan.xie@mbzuai.ac.ae}

\usepackage{bibentry}

\begin{document}

\maketitle

\begin{abstract}
Feedback-driven iteration is widely used to improve LLM-generated scientific
ideas, but existing systems either leave revision decisions implicit in
generation, align revisions with feedback-localized weaknesses, or restrict
action selection to system-specific operations. This is problematic because
problems, hypotheses, and solutions are interdependent: the location where a
weakness manifests does not itself determine the revision scope that will
produce a better successor. We term this phenomenon diagnosis--intervention mismatch.
We introduce
\textbf{\textit{P}}redictive
\textbf{\textit{R}}outing of
\textbf{\textit{I}}nterventions by
\textbf{\textit{M}}odeling
\textbf{\textit{E}}ffects, \textbf{\textit{PRIME}}, a plug-and-play feedback-to-revision decision layer
for existing scientific ideators. PRIME operates in three stages:
(1) \textit{Structural Attribution} interprets feedback over the
problem--hypothesis--solution structure without fixing a revision target;
(2) \textit{Counterfactual Routing} constructs alternative scope-conditioned
interventions and compares their projected successor utility and risks through
lightweight rollouts; and
(3) \textit{Intervention Execution} compiles the selected intervention into an
ordered generator-facing plan with semantic actions, preservation constraints,
and dependency requirements for successor generation.
We validate diagnosis intervention mismatch on 112 human-annotated idea--feedback pairs. Across four benchmarks, two ideators, and fixed- and open-problem settings, PRIME improves successor quality and routing reliability over strong baselines; removing counterfactual routing reduces quality from 3.463 to 1.960 and increases route regret from 0.426 to 1.929, confirming cross-scope effect modeling as the primary source of its gains.
\end{abstract}


\section{Introduction}
With the widespread adoption of large language models in scientific research automation, LLM-based scientific idea generation has emerged as an important research direction
\cite{radensky2026humanllmcompoundscientificideation,
luo2025llm4srsurveylargelanguage,shahhosseini2025large,hasib2025process,
reddy2025towardsscientific,radensky2024scideator,
chen2025automatedinterpretable,
gelles2024resourcedemocratization,
xiong2024improvingscientifichypothesisgeneration,
mishra2026ainsteinllmssolveresearch,
kon2025exp,
liu2025hypobench}.
Within this area, feedback-conditioned scientific idea revision supports the generation of successor ideas based on evaluations of a current idea.
For example, in idea refinement, feedback is used to address the limitations of the current idea and improve its hypotheses, methodology, and validation plan
\cite{deshwal2025adaptiveexperimental,madaan2023self,alkan2025surveyhypothesisgenerationscientific,liu2025personaflow,keya2025sciideacontextawarescientificideation};
in idea evolution, feedback can guide the system in adjusting its research trajectory and exploring alternative hypotheses, solutions, or problem formulations
\cite{yuksekgonul2026learning,zhang2025knowledgedriven,kulkarni2025scientific,nigam2024accelerontool}.
Despite their different settings, these scenarios share the same core objective: translating feedback into concrete updates to successor ideas.
Reliable feedback also depends on judge quality: recent evidence shows that LLM judges can favor presentation style over scientific substance when evaluating research ideas~\cite{ji2026stylewins}. At a later stage of scientific automation, LabGuard grounds natural-language laboratory rules into runtime guards for embodied laboratory agents~\cite{yang2026labguard}; PRIME addresses the earlier step of revising ideas in response to feedback.

\begin{table}[t]
    \centering
    \scriptsize
    \setlength{\tabcolsep}{1.7pt}
    \renewcommand{\arraystretch}{0.5}
    \resizebox{\columnwidth}{!}{%
        \begin{tabular}{@{}lccccccc@{}}
            \toprule
            \textbf{Method}
            & \makecell{\textbf{Feedback} \\ \textbf{update}}
            & \makecell{\textbf{Explicit} \\ \textbf{action}}
            & \makecell{\textbf{Editable} \\ \textbf{state}}
            & \makecell{\textbf{Output} \\ \textbf{search}}
            & \makecell{\textbf{Scope-effect} \\ \textbf{routing}}
            & \makecell{\textbf{Plug-in} \\ \textbf{layer}}
            & \makecell{\textbf{Mismatch} \\ \textbf{audit}}
            \\
            \midrule
            Self-Refine / Reflexion
            & $\checkmark$
            & $\times$
            & $\times$
            & $\times$
            & $\times$
            & $\times$
            & $\times$
            \\
            ResearchAgent / EvoIdeator
            & $\checkmark$
            & $\checkmark$
            & $\times$
            & $\times$
            & $\times$
            & $\times$
            & $\times$
            \\

            IdeaSynth
            & $\checkmark$
            & $\checkmark$
            & $\checkmark$
            & $\times$
            & $\times$
            & $\times$
            & $\times$
            \\

            Deep Ideation / VirSci
            & $\checkmark$
            & $\checkmark$
            & $\times$
            & $\times$
            & $\times$
            & $\times$
            & $\times$
            \\

            AI Scientist / Co-Scientist
            & $\checkmark$
            & $\times$
            & $\times$
            & $\checkmark$
            & $\times$
            & $\times$
            & $\times$
            \\

            Nova / Chain of Ideas
            & $\checkmark$
            & $\times$
            & $\times$
            & $\checkmark$
            & $\times$
            & $\times$
            & $\times$
            \\

            EIG
            & $\checkmark$
            & $\checkmark$
            & $\checkmark$
            & $\times$
            & $\times$
            & $\times$
            & $\times$
            \\

            \textbf{PRIME}
            & $\checkmark$
            & $\checkmark$
            & $\checkmark$
            & $\times$
            & $\checkmark$
            & $\checkmark$
            & $\checkmark$
            \\

            \bottomrule
        \end{tabular}%
    }
\caption{
    Positioning PRIME relative to output-search and feedback-driven revision
    methods. Output-level search selects among generated ideas or trajectories,
    whereas PRIME selects among alternative revisions for the same
    idea--feedback pair before final generation.
}
\label{tab:related_work_positioning}
\end{table}

\begin{figure*}[t]
    \centering
    \includegraphics[width=\textwidth]{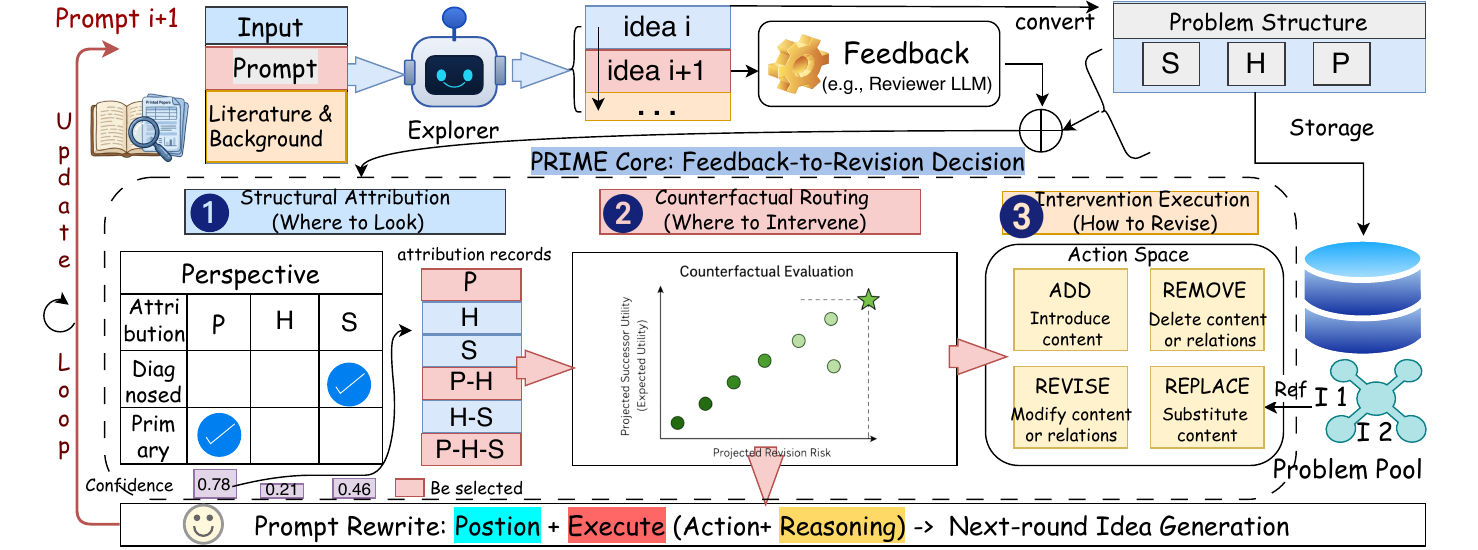}
    \caption{Position–Execute Iterative Refinement. PRIME structures each idea into problem (P), hypothesis (H), and solution (S), positions feedback across their components and relations, selects an action, remove, revise, or replace; rewrites the prompt with explicit reasoning to iteratively generate improved ideas.}
    \label{fig:figure2}
\end{figure*}

Based on how feedback is translated into subsequent revision actions, existing systems can be broadly categorized into three types. The first is holistic rewriting, exemplified by Self-Refine~\cite{madaan2023self}, in which the current idea, together with an overall score, natural-language critique, or reviewer feedback, is provided to a generative model to produce a complete successor version. In this approach, both what should be revised and the appropriate scope of revision remain implicit in the rewriting process. The second is localized corrective revision, exemplified by ResearchAgent~\cite{baek2025researchagent} and EvoIdeator~\cite{sauter2026evoideator}, in which feedback is first grounded in specific text spans, evaluation dimensions, or idea components, after which localized corrections are generated for the diagnosed weaknesses. The third is workflow-coupled revision, exemplified by Deep Ideation~\cite{zhao2025deep}, VirSci~\cite{su2025many}, and EIG~\cite{dong2026evolvingideagraphslearnable}, in which the next action is selected from a predefined set of revision operations according to the current idea's evaluation scores, defect categories, or structured state, using either predefined rules or a learned controller. These three categories respectively leave revision decisions implicit in holistic rewriting, bind them directly to the weaknesses localized by feedback, or constrain them to an operation space predefined by a particular system.

Table~\ref{tab:related_work_positioning} situates these revision mechanisms within the broader landscape of scientific ideation methods.
The first category leaves the revision decision implicit in generation, while the second commonly treats the diagnosed weakness as the direct target of revision~\cite{madaan2023self,baek2025researchagent,sauter2026evoideator}. 
The third makes action selection explicit but couples the decision to a system-specific state representation and a predefined operation space~\cite{dong2026evolvingideagraphslearnable}. None of these mechanisms provides a general decision layer for translating feedback into revision scopes that can be executed by existing idea generators. This limitation is particularly important in scientific ideation, where the problem, hypothesis, and solution are interdependent~\cite{xiong2024kgcoi,pu2025ideasynth,dong2026evolvingideagraphslearnable}. The same observable weakness may arise from causes at different levels, while interventions at different levels may propagate through these dependencies in different ways.
Taken together, these methods leave a fundamental feedback-to-revision gap unresolved: "where a weakness is diagnosed does not by itself determine where revision should be applied".

Bridging this gap presents three key challenges.
First, the locus where feedback manifests may not be the component that actually causes the problem. The same feedback may correspond to multiple plausible intervention loci, making the revision scope difficult to determine uniquely~\cite{xiong2024kgcoi,pu2025ideasynth,dong2026evolvingideagraphslearnable}.
Second, the appropriate intervention scope cannot be directly observed before revision. For a given idea, only the successor produced by the selected revision is typically observed, while exhaustively executing all alternatives is prohibitively expensive~\cite{hu2024nova,wang2026flowpie,sauter2026evoideator,li2026evogens}.
Third, translating the intervention scope into actionable instructions that guide the generator to revise target components, update dependencies, and preserve valid content~\cite{baek2025researchagent,pu2025ideasynth,dong2026evolvingideagraphslearnable}.

We annotate 112 human-annotated idea--feedback pairs, and evaluate six baselines, revealing a diagnosis--intervention mismatch in scientific ideation: feedback may target a different scope from the one that produces a higher-quality successor idea.
To bridge this gap, as shown in Fig.~\ref{fig:figure2}, we introduce \textbf{\textit{P}}redictive \textbf{\textit{R}}outing of \textbf{\textit{I}}nterventions by \textbf{\textit{M}}odeling \textbf{\textit{E}}ffects, \textbf{\textit{PRIME}}, a plug-and-play feedback-to-revision decision layer that can be inserted between an existing feedback provider and idea generator. PRIME consists of three components:
(i) \textit{Structural Attribution}, which interprets feedback over the research problem, scientific hypothesis, solution, and their dependencies to distinguish where a weakness manifests from its potential structural sources;
(ii) \textit{Counterfactual Routing}, which constructs candidate interventions covering the feedback-localized component, its potential upstream causes, cross-component dependencies, and problem-reframing directions, and uses lightweight successor rollouts to evaluate their feedback resolution, prospective utility, and structural risks before selecting an intervention scope; and
(iii) \textit{Intervention Execution}, which translates the selected intervention into an ordered, generator-facing revision plan specifying the target content, concrete semantic operations, preservation constraints, and required dependency updates.
Through these three components, PRIME invokes the original idea generator to produce a complete successor idea.
We evaluate PRIME on four benchmarks. Compared with direct feedback updating, PRIME-Refinement and PRIME-Evolution improve 14/16 and 15/16 metrics, respectively, and jointly achieve the best results on 10/16 metrics, showing that effective revision requires deciding not only how to address feedback, but also where to intervene.
\begin{figure*}[t]
    \centering
    \includegraphics[width=\textwidth]{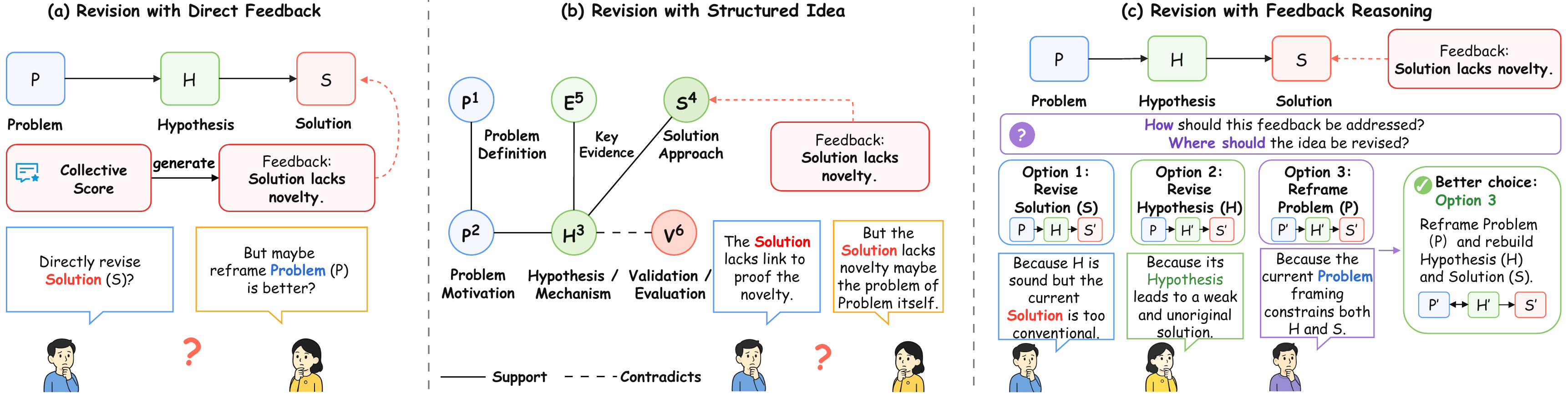}
    \caption{
    Feedback-guided revision through structured reasoning.
    (a) Holistic and localized revision: decisions are implicit in generation or tied to feedback-localized weaknesses.
    (b) Workflow-coupled revision: actions are selected from a system-specific operation space.
    (c) Feedback-reasoned revision: diagnosis revision mismatch is explicitly modeled to guide revision.
    }
    \label{fig:figure3}
\end{figure*}

Our main contributions and findings are:
(i) We formulate the diagnosis--intervention mismatch in scientific ideation, where a feedback-localized weakness does not uniquely determine the revision scope, and validate it through a controlled audit of 114 human-annotated idea--feedback pairs;
(ii) We propose PRIME, a plug-and-play feedback-to-revision decision layer that performs revision planning through Structural Attribution, Counterfactual Routing, and Intervention Execution without replacing the underlying ideator;
(iii) Across four benchmarks, two ideators, and both refinement and evolution settings, PRIME jointly achieves the best results on 20/32 metric ideator combinations; in the intervention audit, PRIME-Evolution further achieves the highest successor utility of 64.16 and the lowest route regret of 3.93.

\section{Related Works}

\paragraph{Iterative Scientific Ideation and Direct Revision.}
As shown in Fig.~\ref{fig:figure3}(a,b), Iterative scientific ideation systems improve candidate ideas through repeated critique, review, refinement, or evolution. General LLM self-improvement methods, such as Self-Refine and Reflexion, show that verbal feedback can drive iterative revision \cite{shinn2023reflexion}. In scientific ideation, ResearchAgent, IdeaSynth, AI Scientist, Co-Scientist, Chain of Ideas, Many Heads, and SciDeator employ reviewing agents, literature-grounded critiques, rubric-based judgments, or multi-agent interactions to improve generated proposals \cite{baek2025researchagent,lu2024aiscientistfullyautomated,Gottweis_2026,li2025chainofideas,su2025headsbetteroneimproved}. 
These systems demonstrate the value of evaluation-driven iteration, but they use evaluation as a direct revision signal, for example, by rewriting the entire proposal, revising its weakest aspect, or continuing along the current trajectory. 
PRIME explicitly models the diagnosis-to-intervention decision instead of treating the diagnosed locus as the revision target.

\paragraph{Structured State and Graph-Based Edit Control.}
Structured ideation methods reduce the opacity of text-only revision by exposing intermediate idea states. SciPIP, KG-CoI, Graph2Idea, and IdeaSynth ground ideation in paper graphs, knowledge graphs, graph-structured contexts, or decomposed idea facets \cite{wang2024scipip,xiong2024kgcoi,li2026graph2idearetrievalaugmentedscientificideageneration,pu2025ideasynth}. Most closely related to our setting, EIG represents an evolving proposal as an idea graph and learns edit-and-commit operations over localized graph structures \cite{dong2026evolvingideagraphslearnable}. This line of work shows that structured weakness localization and editable state representations are useful. 
EIG controls localized edits within its idea-graph workflow, whereas PRIME selects an intervention scope without equating the diagnosed locus with the revision target.

\paragraph{Search and Evolution over Idea Spaces.}
Search- and evolution-based methods broaden scientific ideation by exploring multiple candidate ideas, branches, or trajectories. Nova, Chain of Ideas, AI Scientist, Co-Scientist, FlowPIE, EvoIdeator, and EvoGens\cite{jin2026toward,jiang2025aide,yamada2025ai,chen2026mars,du2026datamaster,liu2026adaptive,yao2023tree} improve idea generation through planning, search, population evolution, checklist-guided optimization, or test-time exploration \cite{hu2024nova,lu2024aiscientistfullyautomated,Gottweis_2026,wang2026flowpie,sauter2026evoideator,li2026evogens}. Their selection targets are typically generated ideas, branches, population members, or trajectories. These methods do not explicitly ask whether, given the same feedback-localized weakness, the next update should take the form of a local refinement, an upstream rationale revision, or a broader problem-level evolution. 
PRIME instead selects a semantic intervention scope before successor generation, turning feedback into an explicit revision decision for the existing ideator.

\section{Problem Formulation}
\label{sec:problem-formulation}
We denote the research context, scientific idea generator, and current idea by
\(C\), \(G\), and \(I_t\), respectively. Given evaluation criteria
\(\Gamma\), an evaluator produces feedback
\(f_t = J(I_t, C, \Gamma)\). Rather than using \(f_t\) directly as an edit
instruction, we study how to select a revision by jointly considering the
feedback and the structural context of the current idea.

The feasible intervention set is
\(\mathcal{A}_t = \mathcal{A}(I_t, f_t, C)\). Each
\(a \in \mathcal{A}_t\) specifies a revision scope \(s(a)\) and a corresponding
revision, inducing a potential successor
\(I_{t+1}^{a} \sim G(\,\cdot \mid C, I_t, f_t, a)\). The
feedback-to-revision decision is
\begin{equation}
    \resizebox{0.9\linewidth}{!}{$
        \displaystyle
        a_t^{*}
        \in
        \operatorname*{arg\,max}_{a \in \mathcal{A}(I_t, f_t, C)}
        \mathbb{E}_{I_{t+1}^{a}}
        \left[
            U(I_{t+1}^{a}; f_t, C, \Gamma)
            -
            \lambda \mathcal{R}(I_{t+1}^{a}, I_t, a)
        \right]
    $}
    \label{eq:feedback-to-revision-objective}
\end{equation}

where \(U\) measures feedback resolution and scientific quality, while
\(\mathcal{R}\) captures revision risks such as loss of valid content,
dependency inconsistency, and task drift. Conditioning the intervention set on
both \(I_t\) and \(f_t\) captures that the same feedback may admit different
revisions under different idea structures, while the expectation reflects that
their successor effects are unobserved before execution.

The feedback-localized locus is
\(\ell_t = \operatorname{Loc}(I_t, f_t)\), and a diagnosis-aligned intervention
\(a_t^{\mathrm{diag}}\) satisfies
\(s(a_t^{\mathrm{diag}}) = \ell_t\). Diagnosis--intervention mismatch occurs
when an intervention with \(s(a) \neq \ell_t\) achieves a higher objective
value than \(a_t^{\mathrm{diag}}\). Thus, the diagnosed locus informs
intervention selection but does not by itself determine the revision scope.

\section{PRIME}
\subsection{Overview}
PRIME approximates the feedback-to-revision decision in Equation.~\ref{eq:feedback-to-revision-objective} through three stages. \textit{Structural Attribution} interprets feedback over the problem, hypothesis, solution, and their dependencies without fixing a revision target; \textit{Counterfactual Routing} compares alternative intervention scopes using their projected successor effects; and \textit{Intervention Execution} translates the selected intervention into an ordered revision plan for the original generator. PRIME therefore operates as a plug-in decision layer without modifying the feedback provider or idea generator.

\subsection{Structural Attribution}
\label{subsec:structural-attribution}
Feedback identifies an observable weakness but does not uniquely reveal its
structural source. Structural Attribution therefore interprets \(f_t\) over
the structure of \(I_t\) without selecting an intervention scope. PRIME normalizes \(f_t\) into \(\bar{f}_t\), which contains an improvement goal,
supporting evidence, preservation requirements, and revision constraints. The
current idea is then represented as
\begin{equation}
    \resizebox{0.9\linewidth}{!}{$
        \displaystyle
        X_t
        =
        \operatorname{Structure}(I_t)
        =
        (P_t, H_t, S_t),
        \qquad
        P_t \rightarrow H_t \rightarrow S_t.
    $}
    \label{eq:structural-representation}
\end{equation}

where \(P_t\), \(H_t\), and \(S_t\) denote the research problem, scientific
hypothesis, and solution, respectively. The dependencies require the hypothesis
to address the problem and the solution to operationalize or test the
hypothesis.

PRIME interprets \(\bar{f}_t\) through six structural views
\(\mathcal{U} = \{P, H, S, PH, HS, PHS\}\). The first three views examine
individual components, \(PH\) and \(HS\) examine adjacent dependencies, and
\(PHS\) examines the complete reasoning chain. For each
\(u \in \mathcal{U}\), PRIME produces one attribution record containing
supporting evidence, the observed failure, a plausible structural explanation,
the corresponding repair requirement, downstream consequences, severity, and
confidence. The module output is
\begin{equation}
    \mathcal{Z}_t
    =
    \operatorname{Attribute}(X_t, \bar{f}_t, C)
    =
    \left\{
        z_t^u \mid u \in \mathcal{U}
    \right\}.
    \label{eq:structural-attribution}
\end{equation}

The records in \(\mathcal{Z}_t\) provide diagnostic evidence rather than
revision actions: they describe which structural conditions are deficient
without prescribing which components should be modified. Consequently, one
attribution may support multiple intervention scopes in Counterfactual Routing.

\subsection{Counterfactual Routing}
\label{subsec:counterfactual-routing}

Structural Attribution provides evidence for revision but does not determine
which intervention will produce a better successor. Counterfactual Routing
therefore constructs candidate interventions and compares their projected
effects before selecting one for execution.

\paragraph{Candidate Intervention Construction.}
An intervention separates where to revise from how to revise. Its scope
satisfies
\(s(a) \in \mathcal{M} = \{P, H, S, PH, HS, PHS\}\), where each scope is an
edit mask. The \(P\) mask covers the problem framing, scope, bottleneck, and
objective; \(H\) covers the hypothesis and its mechanism; and \(S\) covers the
solution, its mechanism, and validation. Composite masks permit joint revisions
to the corresponding components, while fields outside the selected mask are
preserved. Dependencies are represented through explicit requirements and risks
rather than automatic dependency closure.

The revision within a scope is specified by an ordered action plan
\(\Pi_t^a\) composed of \textsc{Add}, \textsc{Remove}, \textsc{Revise}, and
\textsc{Replace} operations. Each step identifies its target component and
field, concrete instruction, success condition, and dependencies on preceding
steps. \textsc{Add} introduces missing content, \textsc{Remove} deletes
unsupported content, \textsc{Revise} modifies an existing commitment while
retaining its role, and \textsc{Replace} substitutes it with a new commitment.

For interventions containing \(P\), the problem operation is
\textsc{Update} or \textsc{Switch}. \textsc{Update} changes the problem framing
while preserving its identity, whereas \textsc{Switch} moves to another
context-grounded problem and is available only in evolution settings.
Interventions not containing \(P\) use \textsc{None}.

\paragraph{Problem-Reference Conditioning.}
Interventions involving \(P\) are conditioned on a persistent Problem Pool
\(\mathcal{P}_t\), which maintains the active problem identity and grounded
alternative references. A \textsc{P-Update} uses the active reference as an
identity anchor, whereas a \textsc{P-Switch} uses an alternative reference as
its target. The pool only supplies reference conditions for candidate
construction and does not rank interventions.
PRIME-Refinement uses a fixed reference policy and disables
\textsc{P-Switch}; PRIME-Evolution uses an open policy and additionally
constructs \textsc{P-Switch} candidates from grounded alternatives in
\(\mathcal{P}_t\). Both variants otherwise use the same attribution, routing,
and execution procedure. The open policy permits but does not force evolution:
only a selected \textsc{P-Switch} realizes an evolution step.

Using the structural representation \(X_t\), normalized feedback
\(\bar{f}_t\), and attribution \(\mathcal{Z}_t\), PRIME constructs
\begin{equation}
    \widetilde{\mathcal{A}}_t
    =
    \bigcup_{m \in \mathcal{M}}
    \operatorname{Construct}
    \left(
        m,
        X_t,
        \bar{f}_t,
        \mathcal{Z}_t,
        C
    \right)
    \subseteq
    \mathcal{A}_t.
    \label{eq:candidate-construction}
\end{equation}
Each candidate additionally cites supporting attributions
\(\mathcal{Z}_t^a \subseteq \mathcal{Z}_t\) and carries preservation
requirements \(\mathcal{K}_t^a\) and dependency requirements
\(\Lambda_t^a\). The supporting attribution need not have the same label as
\(s(a)\): evidence from one structural view may support a local, upstream, or
joint intervention.

\paragraph{Counterfactual Evaluation and Selection.}
For each candidate \(a \in \widetilde{\mathcal{A}}_t\), PRIME generates a
lightweight structured successor
\begin{equation}
    \widetilde{X}_{t+1}^{a}
    =
    \operatorname{Rollout}
    \left(
        X_t,
        \bar{f}_t,
        C,
        a
    \right).
    \label{eq:counterfactual-rollout}
\end{equation}
The rollout applies the declared actions while copying fields outside their
targets unchanged. It approximates the potential successor
\(I_{t+1}^{a}\) in Equation~\ref{eq:feedback-to-revision-objective}, but is not
the final idea produced by \(G\).

A routing evaluator estimates the prospective utility
\(\widehat{U}_t(a)\) and revision risk
\(\widehat{\mathcal{R}}_t(a)\) of each rollout. The former captures feedback
resolution and scientific quality, while the latter captures unsupported
changes, infeasibility, dependency breakage, task drift, overcomplexity, and
uncertainty. Candidates violating predefined scope, preservation, risk, or
problem-reference constraints are removed. PRIME then selects
\begin{equation}
    \begin{aligned}
        \widehat{V}_t(a)
        &=
        \widehat{U}_t(a)
        -
        \lambda \widehat{\mathcal{R}}_t(a), \\
        \hat{a}_t
        &=
        \operatorname*{arg\,max}_{
            a \in \widetilde{\mathcal{A}}_t^{\mathrm{valid}}
        }
        \widehat{V}_t(a).
    \end{aligned}
    \label{eq:counterfactual-routing}
\end{equation}

If no candidate passes the strict validity constraints, PRIME selects the
highest-ranked executable candidate. Counterfactual Routing outputs the
selected intervention \(\hat{a}_t\) and its projected successor
\(\widetilde{X}_{t+1}^{\hat{a}_t}\) for Intervention Execution.

\subsection{Intervention Execution}
\label{subsec:intervention-execution}
The selected intervention must be translated into instructions that the
original generator can execute without reconsidering its scope. PRIME therefore
compiles \(\hat{a}_t\) and its projected successor into a generator-facing
specification and invokes \(G\) once:
\begin{equation}
    \begin{aligned}
        \chi_{t+1}
        &=
        \operatorname{Compile}
        \left(
            C,
            I_t,
            \bar{f}_t,
            \hat{a}_t,
            \widetilde{X}_{t+1}^{\hat{a}_t}
        \right), \\
        I_{t+1}
        &=
        G\!\left(
            \chi_{t+1}
        \right).
    \end{aligned}
    \label{eq:intervention-execution}
\end{equation}

The execution specification \(\chi_{t+1}\) contains the selected scope
\(s(\hat{a}_t)\), ordered action plan \(\Pi_t^{\hat{a}_t}\), preservation
requirements \(\mathcal{K}_t^{\hat{a}_t}\), and dependency requirements
\(\Lambda_t^{\hat{a}_t}\). The projected successor provides a semantic target
for these actions rather than a draft to be copied.

The generator is instructed to execute the selected actions in order, preserve
content outside \(s(\hat{a}_t)\), satisfy the specified dependencies, and return
a complete successor idea in the same format as \(I_t\). The compiled
specification instructs \(G\) not to introduce additional revision targets from
the raw feedback or to modify content outside the selected scope. Because PRIME
communicates with \(G\) through \(\chi_{t+1}\) without modifying its parameters,
it can be combined with different instruction-conditioned idea generators.

\section{Experiments}
\subsection{Experimental Setup}
\label{subsec:experimental_setup}
\paragraph{Baselines.}
We compare PRIME with six feedback-to-revision strategies.
(1) Flat Feedback directly injects a single natural-language improvement direction into the subsequent generation task. 
(2) Problem-First H/S fixes the initial problem and revises only the hypothesis and solution. 
(3) Problem Regeneration updates only the problem used to guide the next iteration. 
(4) Direct Feedback Decomposition generates an ordered, free-form list of revisions and directly executes the highest-ranked item. 
(5) Feedback Reasoning separates feedback generation from revision planning and serves as a simplified PRIME variant without counterfactual cross-scope comparison.
(6) Graph-Based grounds feedback and localized edits in a structured problem--hypothesis--solution graph, while its reasoning variant explicitly selects a graph edit before generation. 
For each baseline, we evaluate two models: Direct Prompting and SciMon \cite{wang2024scimon}.

\paragraph{Benchmark.}
We evaluate PRIME on four scientific idea benchmarks: AI Idea Bench~\cite{qiu2025ai}, IdeaBench~\cite{guo2025ideabench}, LiveIdeaBench~\cite{ruan2024liveideabench}, and MLR-Bench~\cite{chen2026mlr}. These benchmarks follow two evaluation protocols. The first assesses idea reconstruction by comparing generated ideas with target ideas from real papers. AI Idea Bench uses I2I, I2T, and MCQ to evaluate idea matching, topic alignment, and identification of key research content. IdeaBench further uses the novelty score NIS and feasibility score FIS to assess innovation relative to existing work and practical implementability. The second category follows a paper-review-style protocol, with multiple LLM judges directly assessing overall idea quality. LiveIdeaBench evaluates originality, feasibility, and clarity, focusing on whether an idea forms a complete, coherent, and worthwhile research plan. MLR-Bench evaluates clarity, novelty, feasibility, and significance to determine whether an idea approaches the quality expected of a machine learning research paper. For fair comparison, all methods use the same scientific context, reference materials, feedback provider, Qwen3.5-4B generator, 20 iterations, and a generation budget of 2,048 tokens on each benchmark.
For all experiments, we compute the score for each idea generated across all iterations and then report the average score.


\begin{table*}[t]
\centering
\setlength{\tabcolsep}{1.2pt}
\renewcommand{\arraystretch}{0.45}

\resizebox{\textwidth}{!}{%
\begin{tabular}{@{}ll*{16}{c}@{}}
    \toprule

    \textbf{Update Policy}
    & \textbf{Ideator}
    & \multicolumn{3}{c}{\textbf{AI Idea Bench}}
    & \multicolumn{4}{c}{\textbf{IdeaBench}}
    & \multicolumn{3}{c}{\textbf{LiveIdeaBench}}
    & \multicolumn{6}{c}{\textbf{MLR-Bench}}
    \\
    \cmidrule(lr){3-5}
    \cmidrule(lr){6-9}
    \cmidrule(lr){10-12}
    \cmidrule(lr){13-18}

    &
    & \textbf{I2I}
    & \textbf{I2T}
    & \textbf{IMCQ}
    & \textbf{SemSim}
    & \textbf{Overlap}
    & \textbf{NIS}
    & \textbf{FIS}
    & \textbf{Orig.}
    & \textbf{Feas.}
    & \textbf{Clar.}
    & \textbf{Cons.}
    & \textbf{Clar.}
    & \textbf{Nov.}
    & \textbf{Feas.}
    & \textbf{Sig.}
    & \textbf{Overall}
    \\
    \midrule



    Flat Feedback
    & Direct Prompting
    & 2.656 & 4.690 & 0.744 
    & 0.647 & 1.774 & 0.885 & 0.165
    & 5.506 & 4.810 & 5.112
    & 7.192 & 6.338 & 6.657 & 5.689 & 7.061 & 6.411 
    \\

    & SciMON
    & 2.648 & 4.560 & 0.734 
    & 0.599 & 1.540 & 0.996 & 0.029
    & 6.50 & 2.528 & 3.776
    & 7.839 & 6.533 & 7.676 & 5.113 & 7.673 & 6.561 
    \\
    \midrule

    Problem-First H/S
    & Direct Prompting
    & 2.685 & 4.716 & 0.759 
    & 0.647 & 2.011 & 0.937 & 0.164
    & 6.366 & 4.683 & 5.582
    & 7.734 & 6.657 & 6.905 & 6.237 & 7.567 & 6.896 
    \\

    & SciMON
    & 2.621 & 4.469 & 0.707 
    & 0.642 & 1.660 & 0.997 & 0.017
    & 6.938 & 2.426 & 3.637
    & 7.240 & 6.144 & 7.774 & 4.506 & 7.508 & 6.273 
    \\
    \midrule

    Problem Regeneration
    & Direct Prompting
    & 2.519 & 4.556 & 0.766 
    & 0.647 & 1.699 & 0.864 & 0.186
    & 6.180 & 4.171 & 4.916
    & 7.577 & 6.660 & 7.052 & 5.703 & 7.595 & 6.750 
    \\

    & SciMON
    & 2.163 & 3.799 & 0.684 
    & 0.641 & 1.433 & 0.994 & 0.003
    & 6.581 & 2.282 & 3.287
    & 6.989 & 6.139 & 7.865 & 4.502 & 7.405 & 6.208 
    \\
    \midrule

    Feedback Decomp.
    & Direct Prompting
    & 2.609 & 4.668 & 0.736 
    & 0.649 & 2.029 & 0.799 & 0.238
    & 5.616 & 4.261 & 4.626
    & 7.001 & 6.170 & 6.615 & 5.684 & 7.092 & 6.357 
    \\

    & SciMON
    & 2.550 & 4.257 & 0.717 
    & 0.620 & 1.858 & 0.994 & 0.015
    & 6.798 & 2.170 & 3.501
    & 7.475 & 6.373 & 8.077 & 4.396 & 7.882 & 6.504 
    \\

    \midrule

    Feedback Reasoning
    & Direct Prompting
    & 2.687 & 4.744 & 0.767 
    & 0.648 & 1.873 & 0.922 & 0.528
    & 6.403 & 4.622 & 5.072
    & 7.827 & 6.614 & 7.333 & 5.879 & 7.899 & 7.055
    \\

    & SciMON
    & 2.480 & 4.700 & 0.720 
    & 0.646 & 1.700 & 0.940 & 0.600
    & 6.400 & 4.800 & 5.160
    & 7.640 & 6.400 & 7.400 & 5.800 & 7.80 & 6.900
    \\
    \midrule

    \textbf{PRIME-Refinement}
    & Direct Prompting
    & 3.490 & 4.857 & 0.939 
    &  0.655 & 2.500 & 0.958 & 0.271
    & 6.060 & 3.900 & 4.800
    & 8.180 & 6.540 & 7.040 & 5.880 & 8.060 & 7.100 
    \\

    & SciMON
    & 2.735 & 4.735 & 0.755 
    & 0.650 & 2.082 & 0.980 & 0.163
    & 5.449 & 2.388 & 3.327
    & 7.898 & 6.918 & 7.980 & 4.878 & 8.224 & 6.939 
    \\
    \midrule

    \textbf{PRIME-Evolution}
    & Direct Prompting
    & 3.460 & 4.840 & 0.920 
    & 0.651 & 2.400 & 0.920 & 0.200
    & 6.220 & 4.660 & 5.340
    & 8.320 & 6.820 & 7.020 & 6.140 & 7.960 & 7.260  
    \\

    & SciMON
    & 2.667 & 4.688 & 0.792 
    & 0.657 & 1.918 & 1.000 & 0.143
    & 5.469 & 2.347 & 3.510
    & 8.020 & 7.080 & 7.820 & 5.180 & 8.240 & 7.120 
    \\
    \midrule

    \textbf{Graph-Based}
    & Direct Prompting
    & 3.184 & 4.953 & 0.884 
    & 0.665 & 2.518 & 0.911 & 0.051
    & 5.193 & 3.683 & 3.306
    & 8.015 & 5.707 & 6.966 & 5.228 & 7.898 & 6.589 
    \\

    & w Reasoning 
    & 3.377 & 4.868 & 0.866 
    & 0.660 & 2.337 & 0.852 & 0.264
    & 5.515 & 4.139 & 4.043
    & 8.044 & 6.170 & 7.088 & 5.607 & 7.928 & 6.870 
    \\
    
    & w PRIME-Refinement 
    & 3.440 & 4.940 & 0.840 
    & 0.670 & 2.340 & 0.920 & 0.220
    & 5.940 & 3.940 & 4.720
    & 8.240 & 6.460 & 7.040 & 5.640 & 8.020 & 7.060
\\
    & w PRIME-Evolution 
    & 3.460 & 4.900 & 0.920 
    & 0.648 & 2.300 & 0.840 & 0.320
    & 6.100 & 4.240 & 5.040
    & 8.100 & 6.480 & 6.880 & 5.780 & 7.900 & 7.020 
    \\
    \bottomrule
\end{tabular}%
}
\caption{
 Main results across four scientific idea-generation benchmarks. Rows compare feedback-to-revision policies across two ideators, with the Graph-Based block additionally reporting reasoning and PRIME-enhanced variants. PRIME-Refinement fixes the problem reference, whereas PRIME-Evolution permits context-grounded problem switching. All methods use the same experimental settings within each benchmark. All metrics are higher-is-better.
}
\label{tab:main_results}
\end{table*}

\subsection{Main Results}
The main experiment examines whether explicitly modeling the feedback-to-revision decision improves the quality of iteratively generated successor ideas across ideators and revision settings. Table~\ref{tab:main_results} compares five general update policies with PRIME-Refinement and PRIME-Evolution under Direct Prompting and SciMON across four benchmarks. The Graph-Based block further tests whether PRIME can be integrated with a structured ideator. Using random seed 101, we sample 50 tasks from each benchmark and perform 20 feedback-to-revision iterations for every task under each policy--ideator configuration. All methods share the same Qwen3.5-4B base model, research contexts, feedback provider, generation budget, and benchmark-specific evaluation protocol. Across the five general policies and two PRIME variants, PRIME achieves the best results on 10/16 metrics under both Direct Prompting and SciMON. Compared with Flat Feedback, PRIME-Refinement and PRIME-Evolution improve 14/16 and 15/16 metrics under Direct Prompting, and 11/16 and 13/16 under SciMON, respectively. PRIME-Refinement leads all three AI Idea Bench metrics under Direct Prompting, while PRIME-Evolution achieves the highest MLR-Bench overall scores under both ideators (7.260 and 7.120). Although results on LiveIdeaBench are mixed, adding PRIME-Refinement to the Graph-Based ideator improves all LiveIdeaBench and MLR-Bench metrics, increasing the MLR-Bench overall score from 6.589 to 7.060. These results support PRIME as an effective plug-in layer for feedback-to-revision routing.

\textit{Statistical Significance and Multi-Judge Consistency.} To ensure robust evaluation and reduce judge-specific bias, we randomly sample 50 generated ideas from each of the eight result groups, yielding 400 sampled ideas in total. Each idea is independently scored by three LLM judges, resulting in 1,200 evaluation records. We report exact agreement, $\scriptstyle \mathrm{EA}=N^{-1}\sum_i\mathbf{1}\bigl[y_i^{(a)}=y_i^{(b)}\bigr]$; adjacent agreement, $\scriptstyle \mathrm{AA}=N^{-1}\sum_i\mathbf{1}\bigl[\lvert y_i^{(a)}-y_i^{(b)}\rvert\leq1\bigr]$; and the Pearson correlation between judges. The resulting agreement scores, $\mathrm{EA}=68.7\%$, $\mathrm{AA}=78.9\%$, and Pearson $r=0.72$, indicate strong inter-judge consistency.

\begin{table*}[t]
    \centering
    \footnotesize

    \begingroup
    \setlength{\tabcolsep}{0.95pt}
    \renewcommand{\arraystretch}{0.45}

    \begin{tabularx}{
        \textwidth
    }{
        @{}
        l
        *{9}{c}
        >{\centering\arraybackslash}X
        @{}
    }
        \toprule

        \multicolumn{11}{l}{
            \textbf{(a) Diagnostic--intervention location mismatch}
        }
        \\
        \addlinespace[2pt]

        \textbf{Source or Setting}
        & \textbf{$N$}
        & \textbf{$M@0$}
        & \textbf{$M@1$}
        & \textbf{$M@2$ [95\% CI]}
        & \textbf{$M@5$}
        & \textbf{$M_{\mathrm{any}}@2$ [95\% CI]}
        & \textbf{Rep.\ $M@2$ [95\% CI]}
        & \textbf{$R_{\mathrm{diag}}$}
        & \textbf{95\% CI}
        & \textbf{$p$}
        \\

        \midrule

        Human Groundtruth
        & 112
        & 73.2
        & 72.3
        & 65.2 [56.2, 74.1]
        & 41.1
        & 40.2 [31.2, 49.1]
        & 43.8 [34.8, 52.7]
        & $3.79$
        & [$ 2.74$, 4.93]
        & $.0001$
        \\

        Model Attribution
        & 112
        & 73.2
        & 73.2
        & 61.6 [52.7, 70.5]
        & 38.4
        & 42.9 [33.9, 51.8]
        & 25.0 [17.0, 33.0]
        & $1.13$
        & [$-0.06$, 2.42]
        & $.0386$
        \\

        Human 1
        & 112
        & 73.4
        & 73.4
        & 68.8 [59.6, 77.1]
        & 38.5
        & 65.1 [56.0, 74.3]
        & 28.4 [20.2, 37.6]
        & $1.15$
        & [$ 0.21$, 2.15]
        & $.0137$
        \\

        Human 2
        & 112
        & 76.6
        & 76.6
        & 71.2 [62.2, 79.3]
        & 38.7
        & 68.5 [59.5, 77.5]
        & 29.7 [21.6, 38.7]
        & $1.03$
        & [$ 0.05$, 2.02]
        & $.0212$
        \\

        Human–Human Reliability
        & 77
        & 67.5
        & 66.2
        & 62.3 [51.9, 72.7]
        & 29.9
        & 62.3 [51.9, 72.8]
        & 20.8 [11.7, 29.9]
        & $0.05$
        & [$-1.01$, 1.14]
        & $.4356$
        \\
        \bottomrule
    \end{tabularx}

    \endgroup
    \begingroup
    \setlength{\tabcolsep}{4.2pt}
    \renewcommand{\arraystretch}{0.5}
    \begin{tabularx}{
        \textwidth
    }{
        @{}
        l
        c
        *{7}{c}
        >{\centering\arraybackslash}X
        @{}
    }
        \toprule

        \multicolumn{10}{l}{
            \textbf{(b) Feedback-to-intervention routing on the 112 complete instances}
        }
        \\
        \addlinespace[1.8pt]
        \textbf{Update Policy}
        & \textbf{$N$}
        & \textbf{Diag. Match}
        & \textbf{Target Hit@2 $\uparrow$}
        & \textbf{RepErr@2 $\downarrow$}
        & \textbf{Benefit@2 $\uparrow$}
        & \textbf{Harm@2 $\downarrow$}
        & \textbf{$V_{\mathrm{succ}} \uparrow$}
        & \textbf{$\Delta V_{\mathrm{diag}} \uparrow$}
        & \textbf{$R_{\mathrm{route}} \downarrow$}
        \\

        \midrule

        Flat Feedback
        & 112
        & 23.2
        & 37.5
        & 19.6
        & 57.1
        & 8.9
        & 63.83
        & 3.61
        & 4.25
        \\

        Problem-First H/S
        & 112
        & 16.1
        & 37.5
        & 22.3
        & 56.2
        & 10.7
        & 63.81
        & 3.58
        & 4.27
        \\

        Problem Regeneration
        & 112
        & 22.3
        & 24.1
        & 43.8
        & 28.6
        & 28.6
        & 60.69
        & 0.46
        & 7.40
        \\

        Feedback Decomp.
        & 112
        & 11.6
        & 34.8
        & 25.0
        & 53.6
        & 13.4
        & 63.47
        & 3.24
        & 4.62
        \\

        Feedback Reasoning
        & 112
        & 16.1
        & 38.4
        & 24.1
        & 55.4
        & 12.5
        & 63.60
        & 3.37
        & 4.49
        \\

        PRIME-Refinement
        & 112
        & 45.5
        & 36.6
        & 25.0
        & 50.0
        & 12.5
        & 62.52
        & 2.29
        & 5.56
        \\

        PRIME-Evolution
        & 112
        & \textbf{77.7}
        & \textbf{43.8}
        & \textbf{16.1}
        & \textbf{65.2}
        & \textbf{2.7}
        & \textbf{64.16}
        & \textbf{3.93}
        & \textbf{3.93}
        \\
        \bottomrule
    \end{tabularx}
    \endgroup
    \caption{
        Diagnosis intervention location mismatch and
        feedback-to-intervention routing.
        Panel (a) establishes whether the diagnostic location identifies the
        most effective intervention location.
        Panel (b) evaluates how six update policies convert the same feedback
        into intervention actions on the 112 complete annotated instances.
    }
    \label{tab:dilm_and_policy_routing}
\end{table*}

\subsection{Diagnostic Intervention Mismatch}
We test whether the locus identified by feedback reliably determines the
intervention scope that produces a better successor idea. From 114
human-annotated idea--feedback pairs, we retain 112 complete instances from
IdeaBench, LiveIdeaBench, and MLR-Bench, comprising 42, 42, and 28 instances,
respectively. For each instance, we hold the idea, feedback, context, generator,
and generation budget fixed, and vary only the intervention scope over
\(\{P,H,S,PH,HS,PHS\}\), yielding 672 scope-conditioned outcomes. Human
annotators identify the diagnostic locus using only the original idea and
feedback, without observing successors or policy decisions, while blind
evaluators independently assess the resulting successors. A selection split
defines the \(\epsilon\)-near-optimal intervention set, and a separate
evaluation split measures held-out mismatch and regret. Panel~(a) audits whether
diagnosis identifies a near-optimal intervention, while Panel~(b) applies seven
update policies to the same outcome bank to evaluate feedback-to-intervention
routing.

Table~\ref{tab:dilm_and_policy_routing} shows a substantial
diagnosis--intervention mismatch. At the primary threshold \(\epsilon=2\), the
human-ground-truth diagnosis falls outside the near-optimal intervention set in
\(65.2\%\) of instances; the mismatch remains \(40.2\%\) when all admissible
diagnostic loci are considered and replicates on the held-out split in
\(43.8\%\) of instances. Directly following the diagnosed locus incurs an
average held-out regret of \(3.79\) points
(\(95\%\) CI: \([2.74, 4.93]\), \(p=.0001\)), with similar mismatch rates across
model and individual-human diagnoses (\(61.6\%\)--\(71.2\%\)). For routing, PRIME-Evolution achieves the highest Target Hit@2
(\(43.8\%\)), Benefit@2 (\(65.2\%\)), and successor utility (\(64.16\)), while
obtaining the lowest RepErr@2 (\(16.1\%\)), Harm@2 (\(2.7\%\)), and evaluation
regret (\(3.93\)). PRIME-Refinement does not exhibit the same advantage,
suggesting that restricting problem-level alternatives can be limiting when
broader intervention is beneficial. Overall, the results show that feedback
localization alone is insufficient and that comparing the projected effects of
alternative intervention scopes improves revision routing.

\subsection{Base-Model and Ideators Generalization}
\label{subsec:generalization} 
We evaluate whether PRIME generalizes across both base models and ideation workflows through two controlled substitutions on AI Idea Bench. First, we replace the primary base model with Llama while retaining Direct Prompting as the ideator. Second, we retain the primary base model but replace Direct Prompting with CoI-Agent. All other settings, including the update policies, feedback provider, generation budget, and evaluation protocol, remain unchanged. Table~\ref{tab:model_generator_generalization} reports the results. Under the Llama substitution, PRIME obtains the best I2I, I2T, and IMCQ scores across its refinement and evolution variants. Under the CoI-Agent substitution, PRIME-Evolution achieves the best IMCQ and a near-best I2T score, although Flat Feedback remains stronger on I2I. These results show that PRIME's gains transfer strongly across base models, while its benefits under ideator substitution are metric-dependent.

  \begin{table}[t]
    \centering
    \small
    \setlength{\tabcolsep}{4pt}
    \renewcommand{\arraystretch}{0.45}
    \begin{tabular}{@{}lcccc@{}}
        \toprule
        Update Policy
        & I2I $\uparrow$
        & I2T $\uparrow$
        & IMCQ $\uparrow$
        \\
        \midrule
        \multicolumn{5}{@{}l}{
            \textit{(a) Base-model transfer:
            Llama with Direct Prompting}
        }
        \\
        \cmidrule(lr){1-5}

        Flat Feedback
        & 3.114 & 4.374 & 0.787 
        \\

        Problem-First H/S
        & 3.094 & 4.231 & 0.821 
        \\

        Problem Regeneration
        & 2.876 & 4.238 & 0.816 
        \\
        
        Direct Feedback Decomp.
        & 3.122 & 4.415 & 0.821 
        \\
        
        Feedback Reasoning 
        & 3.573 & 4.568 & 0.883 
        \\
        
        \textbf{PRIME-Refinement}
        & 3.740 & 4.700 & 0.900 
        \\

        \textbf{PRIME-Evolution}
        & 3.740 & 4.720 & 0.860 
        \\
        
        \multicolumn{5}{@{}l}{
            \textit{(b) Ideator transfer:
            CoI-Agent with the original base model}
        }
        \\
        \cmidrule(lr){1-5}

        Flat Feedback
        & 3.402 & 4.964 & 0.840 
        \\

        Problem-First H/S
        & 3.254 & 4.897 & 0.811 
        \\

        Problem Regeneration
        & 2.792 & 4.684 & 0.790 
        \\

        Direct Feedback Decomp.
        & 3.386 & 4.809 & 0.861 
        \\
        Feedback Reasoning 
        & 3.118 & 4.984 & 0.862
        \\
        \textbf{PRIME-Refinement}
        & 3.250 & 4.917 & 0.854 
        \\

        \textbf{PRIME-Evolution}
        & 3.286 & 4.980 & 0.898 
        \\
        
        \bottomrule
    \end{tabular}
\caption{
    Generalization results on AI Idea Bench. Panel~(a) replaces the base model with Llama under Direct Prompting, while Panel~(b) replaces Direct Prompting with CoI-Agent using the original base model.
}
  \label{tab:model_generator_generalization}
  \end{table}

\subsection{Ablation}
\label{subsec:ablation}
  We conduct ablation studies to examine the key design choices of PRIME by removing counterfactual routing, upstream
  intervention candidates, problem-level candidates, and the risk penalty from the full model. As reported in
  Table~\ref{ablation_results}, removing counterfactual routing replaces successor-effect comparison with diagnosis-
  based intervention selection, removing upstream or problem-level candidates restricts the intervention space, and
  removing the risk penalty selects interventions solely according to their predicted utility. Counterfactual routing
  is the most influential component: removing it decreases Quality from 3.463 to 1.960, increases Regret from 0.426 to
  1.929, and decreases Consistency from 3.173 to 1.773. Excluding upstream and problem-level candidates also
  consistently reduces successor quality and increases routing regret. In contrast, removing the risk
  penalty results in only a minor performance degradation, suggesting that projected successor utility dominates
  routing decisions on this benchmark, while risk modeling primarily provides a modest regularization effect.
\begin{table}[t]
\centering
\setlength{\tabcolsep}{3.5pt}
\renewcommand{\arraystretch}{0.45}
\begin{tabular}{@{}lccc@{}}
    \toprule
    \textbf{Variant}
    & \textbf{Quality $\uparrow$}
    & \textbf{Regret $\downarrow$}
    & \textbf{Cons. $\uparrow$}
    \\
    \midrule

    \textbf{PRIME}
    & 3.463 & 0.426 & 3.173
    \\

    w/o Counterfactual Routing
    & 1.960 & 1.929 & 1.773
    \\

    w/o Dependency Closure
    & 3.490 & 0.399 & 3.211
    \\

    w/o Upstream Candidates
    & 3.177 & 0.712 & 2.909
    \\

    w/o Problem Candidates
    & 3.110 & 0.779 & 2.831
    \\

    w/o Risk Penalty
    & 3.459 & 0.430 & 3.158
    \\

    \bottomrule
\end{tabular}
  \caption{Ablation results for PRIME. Quality, Regret, and Cons. denote successor utility, regret relative to the
  post-hoc best intervention, and structural consistency, respectively. Higher Quality and Cons. and lower Regret
  indicate better performance.}
\label{ablation_results}
\end{table}

\section{Conclusion}
\label{sec:conclusion}
In this paper, we identify reference-frame ambiguity as a key challenge in
context-driven scientific idea generation, where ideas may target different
problems and flat comparison may conflate problem selection with solution
quality. We propose PRIME, a plug-and-play framework that recovers latent problem
frames, organizes ideas in a dynamic problem tree, and provides
problem-conditioned evaluation and feedback. Our results show that feedback
localization alone is insufficient for selecting revision scope; by making both
problem positioning and intervention routing explicit, PRIME enables more
reliable comparison, actionable feedback, and controllable iterative improvement
for scientific ideation.

\bibliography{aaai2027}

\end{document}